\pdfoutput=1

\PassOptionsToPackage{dvipsnames}{xcolor}

\documentclass[11pt]{article}

\usepackage[]{./template/acl}

\usepackage{times}
\usepackage{latexsym}

\usepackage[T1]{fontenc}

\usepackage[utf8]{inputenc}

\usepackage{microtype}

\usepackage{stfloats}

\usepackage{ctable} 
\usepackage{subcaption}
\usepackage{amsmath,amsfonts,amssymb}
\usepackage{multicol}
\newsavebox{\algleft}
\newsavebox{\algright}
\usepackage{bm}
\usepackage[linesnumbered,ruled,vlined]{algorithm2e}
\usepackage{hyperref}       
\usepackage{url}            
\usepackage{booktabs}       
\usepackage{amsfonts}       
\usepackage{nicefrac}       
\usepackage{xcolor}         
\usepackage{multirow}
\usepackage{soul}

\graphicspath{{./figure/}}

\title{Scalable and Robust Self-Learning for Skill Routing in Large-Scale Conversational AI Systems}

\author{
  Mohammad Kachuee, Jinseok Nam, Sarthak Ahuja, Jin-Myung Won, Sungjin Lee \\
  Amazon Alexa AI \\
  Seattle, WA \\
  }

\date{}

\begin{document}
\maketitle
\begin{abstract}
Skill routing is an important component in large-scale conversational systems. 
In contrast to traditional rule-based skill routing, state-of-the-art systems use a model-based approach to enable natural conversations.
To provide supervision signal required to train such models, ideas such as human annotation, replication of a rule-based system, relabeling based on user paraphrases, and bandit-based learning were suggested. However, these approaches: (a) do not scale in terms of the number of skills and skill on-boarding, (b) require a very costly expert annotation/rule-design, (c) introduce risks in the user experience with each model update.
In this paper, we present a scalable self-learning approach to explore routing alternatives without causing abrupt policy changes that break the user experience, learn from the user interaction, and incrementally improve the routing via frequent model refreshes. To enable such robust frequent model updates, we suggest a simple and effective approach that ensures controlled policy updates for individual domains, followed by an off-policy evaluation for making deployment decisions without any need for lengthy A/B experimentation.
We conduct various offline and online A/B experiments on a commercial large-scale conversational system to demonstrate the effectiveness of the proposed method in real-world production settings.
\end{abstract}

\section{Introduction}
\begin{figure}[t]
    \centering
        \includegraphics[width=\linewidth]{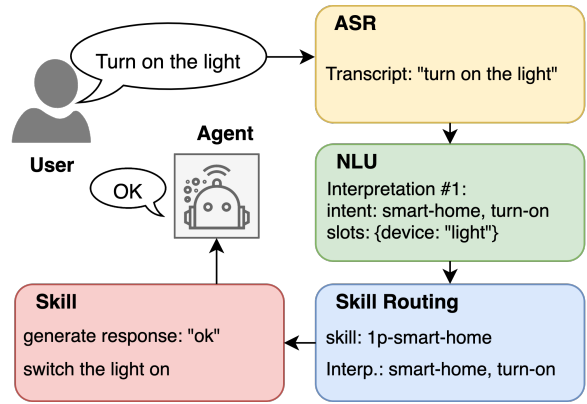}
        \caption{An overview of the major processing steps to handle a user request in a conversational system.}
        \label{fig:conversation_system}
\end{figure}

Large-scale intelligent conversational systems such as Apple Siri, Amazon Alexa, Google Assistant, and Microsoft Cortana are an integral part of the transition from traditional human-machine interactions to seam-less and natural interactions.
A conversational system is a complex interplay of multiple components ranging from the hardware and signal processing blocks to machine learning models.
Figure~\ref{fig:conversation_system} shows an overview of the major processing steps to handle a user request: $(i)$ the automated speech recognition (ASR) block transcribes the utterance along with generating a transcription confidence signal and other voice features such as user's emotion $(ii)$ the natural language understanding (NLU) generates a set of ranked interpretations in terms of user intent as well as named entity resolution and slots corresponding to each interpretation, $(iii)$ a skill routing system uses NLU and ASR outputs as well as other contextual signals to select a skill and NLU interpretation to serve the request, $(iv)$ the selected skill handles the request and generates a response for the user~\citep{sarikaya2017technology}.

To provide the supervision necessary for training skill routing models, different approaches such as replicating a rule-based system, using human annotation, and relabeling based on user paraphrases have been suggested in the literature~\citep{park2020scalable,sarikaya2017technology,sammut2001managing}. Using human annotations is very expensive and suffers from high turn-around times, making it impractical for real-world large-scale systems in which new skills are being introduced frequently. On the other hand, relabeling methods such as the one introduced by \citet{park2020scalable} are limited to cases where we observe enough rephrases with high precision. 

From the scalability and turn-around time perspective, the traditional approach of training models then conducting long A/B experiments before each model deployment results in a limited model update frequency, often insufficient for keeping up with the introduction of new skills and other traffic changes.
An alternative would be to formulate the problem as a contextual bandit and directly aim to maximize the user satisfaction~\citep{karampatziakis2019lessons}. This approach can be more scalable in terms of supervision as user satisfaction is already an established metric in conversational systems~\cite{kachuee2021self}. Also, off-policy evaluation can be used to reduce the need for conducting A/B experiments.
However, in a large-scale production system relying solely on the user satisfaction maximization may cause instabilities due to bandit exploration or even estimation errors in the off-policy learning~\citep{sachdeva2020off,joachims2018deep}.



This paper presents a novel \textit{self-learning} approach based on contextual bandit learning to continuously explore alternative decisions, get user feedback, and learn to improve the skill routing decisions. As frequent model refreshes are a part of the self-learning loop, we suggest a hybrid policy architecture aimed to control policy deviations ensuring consistent and robust improvements to the user experience i.e., not causing an abrupt policy change that results in a broken user experience on certain use cases. The suggested method is simple and yet effective as it supports different levels of robustness-sensitivity for each NLU intent. Furthermore, the proposed approach relies on off-policy evaluation followed by extensive tests rather than the traditional A/B analysis. This approach enables low turn-around time model refreshes in the real service settings, while maintaining the best user experience for business-critical use-cases. To validate the effectiveness of the proposed method, we conduct extensive offline and online A/B experiments on real customer traffic in a real-world large-scale commercial dialogue system.

\section{Related Works}
The first generation of skill routing in conversational systems used a rule-based system to serve a user's request. These rules can be defined at multiple levels and on different signals such as pre-recorded voice, utterance transcript, or NLU interpretation~\citep{sarikaya2017technology,sammut2001managing}. However, rule-based implementations suffer from the inability to generalize and understand natural language variations. Another important drawback of rule-based routing systems is scalability issues arising when dealing with a large number of competing skills and rules~\citep{jadhav2020towards,agostaro2005conversational}.

Model-based conversational systems use machine learning models to understand the user's utterance and select the best skill to serve the request. A model-based system can generalize beyond the capability of a rule-based system as a machine learning model can potentially understand the semantic meaning of a request~\citep{park2020scalable}. Note that despite the promise of better generalization and scalability, in a real-world large-scale system, the transition from a rule-based to a model-based approach is challenging as complex models are known for lack of robustness and interpretability~\citep{li2021neural}.

Providing supervision for model training is an important consideration in training skill routing models. 
A rule-based system can be used to provide a supervision signal to a model, hoping the model to generalize beyond the provided training examples. This kind of replication objective is relatively simple and desirable when considering the robustness aspects; however, in practice, it may not generalize much beyond the rule-based approach~\citep{li2021neural}.

Another line of work is based on relabeling samples by detecting rephrase utterances~\citep{park2020scalable} as users tend to rephrase and repeat when the agent fails to properly respond. However, such a relabeling only covers correction patterns for a subset of traffic presenting only a limited routing improvement opportunity. For example, a user may decide to abandon the dialogue rather than paraphrasing the same request. 


Considering user satisfaction being a major goal of dialogue systems one can use satisfaction as a supervision signal to guide the routing decisions. User satisfaction measurement and prediction in dialogue systems has been studied extensively in the literature~\citep{kachuee2021self,park2020large,bodigutla2019multi,jiang2015automatic}. One possible approach is to formulate the skill routing problem as a contextual bandit problem aiming to maximize the user satisfaction~\citep{karampatziakis2019lessons}. It enables an active exploration of alternative candidates guided by the user experience in user-agent interactions. However, in a real-world production system, it is critical to control the agent behavior changes as excessive exploration or off-policy estimations errors in bandit learning may cause unexpected behavior.



\section{Proposed Method}

\subsection{Problem Definition}
We consider the general problem of skill routing in conversational systems. 
Specifically, we define different pairs of NLU interpretation (e.g., domain, intent, slots, etc.) and skill (e.g., weather skill or shopping skill) as routing candidates, i.e. the action space of our policies.
Given a set of routing candidates and their corresponding contextual signals, encoded in vector space as $X = \{\mathbf{x}_1 \dots \mathbf{x}_T|\mathbf{x}_i \in \mathcal{R}^d \}$, the skill routing agent is tasked to select a routing candidate, $a \in \{1\dots T\}$, to serve the user.

Furthermore, we assume there exists a current, not necessarily optimal, policy denoted by $\Pi_0(a|X)$. The task is to learn from the experiences collected from the current policy interactions in an off-policy setting to train a new policy parameterized by $\theta$, $\Pi_{\theta}(a|X)$, aiming to improve user satisfaction. Here, after taking an action, the agent observes a reward signal, $r$, that is a measure of user satisfaction. The reward signal itself consists of multiple components such as implicit/explicit user feedbacks and machine learning models.

\subsection{Self-Learning Process}
\begin{figure}[t]
    \centering
        \includegraphics[width=\linewidth]{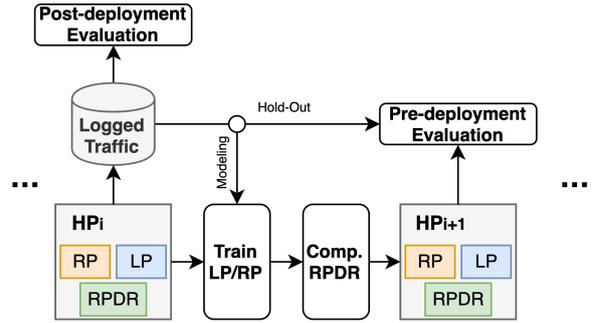}
        \caption{An overview of the self-learning process: model training, RPDR computation, pre-deployment evaluation, and post-deployment evaluation.}
        \label{fig:self_learning_loop}
\end{figure}

Figure~\ref{fig:self_learning_loop} shows an overview of the proposed self-learning process. First, a batch of logged interactions is collected from the current policy (denoted by $HP_i$ in the figure). Then, we use off-policy learning to update the policy using a split of the logged traffic (Section~\ref{sec:model_training} and Section~\ref{sec:hybrid_policy}). The new policy is evaluated before and after the actual deployment enabling the use of guardrail metrics for making a deployment decision as well as tracking the actual online performance of the new model (Section~\ref{sec:pre_post_evaluation}).

\subsection{Model Architecture}
\begin{figure}[t]
    \centering
        \includegraphics[width=0.85\linewidth]{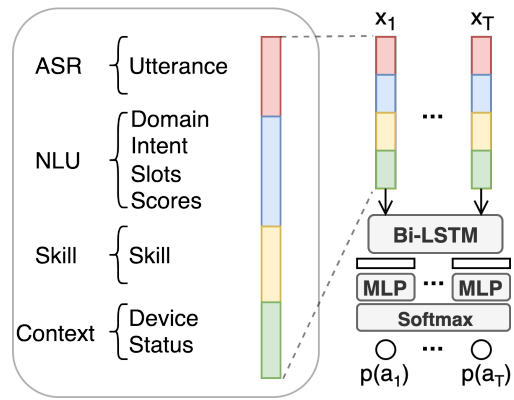}
        \caption{An overview of the proposed model architecture: a set of hypothesis are encoded as vectors ($\mathbf{x}_1 \dots \mathbf{x}_T$) and fed to a bi-directional LSTM which is followed by a shared MLP and a softmax layer to normalize the predicted candidate selection probabilities.}
        \label{fig:model_arch}
\end{figure}

Figure~\ref{fig:model_arch} shows an overview of the model architecture.
Inputs to the model consist of NLU interpretation and skill for each candidate as well as ASR transcription and a diverse set of contextual signals (e.g. ASR confidence, device type, device status, etc.) shared among candidates. 

We encode categorical features using an embedding matrix with a feature size proportional to the square root of the number of unique values. Utterance text is encoded using word vectors and a bi-directional LSTM. The sequence of embedded vectors is reduced via a summation operation, and contextual signals are concatenated to get the final representation i.e. $\mathbf{x} \in \mathcal{R}^d$. Finally, the set of encoded hypotheses, $X$, is sorted based on the NLU interpretation confidence and fed through a bi-directional LSTM, two fully-connected layers, and a softmax activation to output action probabilities for the $\Pi_{\theta}(X)$ policy.

\subsection{Model Training}
\label{sec:model_training}
We define two training objectives: replication policy (RP) and learning policy (LP). RP objective tries to train models that replicate the logged actions. Specifically, we define the RP loss function to minimize:
\begin{equation}
    L_{RP} = \mathbb{E}_{X,a \sim \mathbb{D}} \sum_{i=1}^{T} - \mathbf{1}[a=i] log(\Pi_{\theta}(a|X)).
    \label{eq:loss_rp}
\end{equation}
In short, \eqref{eq:loss_rp} is a cross-entropy loss encouraging the new policy to assign the highest scores to actions that replicate the logged actions. We also explored other alternatives such as KL-divergence or soft-distillation objectives but found that the simple cross-entropy objective is very stable and shows an excellent replication performance.

We define the LP loss function to be an off-policy contextual bandit objective such as the inverse propensity scoring (IPS) objective:
\begin{equation}
    L_{LP} = \mathbb{E}_{X,a,r \sim \mathbb{D}} = r  \frac{\Pi_\theta(a|\mathbf{x})}{\Pi_0(a|\mathbf{x})},
    \label{eq:loss_lp}
\end{equation}
$r$ is the observed reward for taking action $a$ logged in the dataset. The objective of \eqref{eq:loss_lp} trains a policy aimed at maximizing the expected reward. Here, for simplicity, we use the vanilla IPS estimator; however, any other off-policy bandit objective (e.g., doubly-robust estimator) can be used instead.

\subsection{Hybrid Policy}
\label{sec:hybrid_policy}
In a production system, any policy update directly impacts the user experience. Training new policies with a general reward maximization goal, without any control on the changes in behavior, imposes various practical risks. For example, a new model may reduce the quality of skill routing for tail domains while showing a better average performance. As another example, the new policy may explore alternatives aggressively which, considering the turn-around time in the off-policy setting, may cause a widespread negative experience until the next model refresh. To mitigate this issue and limit the changes in the policy behavior in a single model update, we introduce the idea of using a hybrid policy (HP). An HP consists of two internal models trained using the RP and LP objectives. Since the RP replicates the current behavior and the LP tries to maximize the reward, by stochastically selecting RP or LP, we can create a balance between replication of the current behavior and potential improvement in the reward function by making alternative decisions.

\begin{figure}[t]
    \centering
        \includegraphics[width=0.50\linewidth]{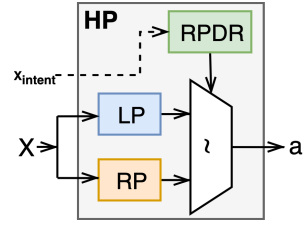}
        \caption{The hybrid policy consists of the LP and RP models as well as the pre-computed RPDR values. At the inference time, the RPDR value corresponding to the NLU top intent used to stochastically decide which model handles that sample.}
        \label{fig:hp_arch}
\end{figure}

Specifically, to create an HP model, we start by training two individual models using the RP and LP objectives. Then, we use the validation set to compute the rate at which LP replicates the current policy for each data segment, computed as:
\begin{equation}
    \kappa_j = \mathbb{E}_{X \sim \mathbb{D}_j} [1 - \frac{|\Pi_{\theta}(X) - \Pi_{0}(X)|_1}{2}] , 
\end{equation}
where $j$ is the index of each data segment and $\kappa_j$ is the expected rate at which the new LP policy replicates the current policy. In this work, we define data segments to be based on the highest scoring NLU intent. Furthermore, we set a desired minimum replication rate ($\Tilde{\kappa}$) for all data segments (e.g. $\Tilde{\kappa}=99\%$). To achieve the desired level of replication, we define the reference policy decision rate (RPDR) as:
\begin{equation}
RPDR_j = 
\begin{cases}
    0 & \text{if } \kappa_j \geq \Tilde{\kappa}\\
    \frac{\Tilde{\kappa} - \kappa_j}{1 - \kappa_j} & \text{otherwise}
\end{cases}
.
\end{equation}
Intuitively, assuming RP is a good replication model, by using RPDR to stochastically decide whether LP or RP should handle each sample, we can achieve the desired level of minimum replication for each segment. The final HP model consists of the LP, RP, and a dictionary of pre-computed RPDR values for each intent. See Figure~\ref{fig:hp_arch} for an illustration of the HP.

To update the HP, depending on the LP/RP update frequency, each time one of the models is trained on the modeling data split, followed by computing the RPDR values (see Figure~\ref{fig:self_learning_loop}). We update LP models more frequently than RP (e.g., LP is updated daily while RP is updated weekly). The reason behind this decision is to limit the changes in the routing behavior for longer periods. The less frequently updated RP model act as a moving average filter, gradually absorbing the LP behavior.

\subsection{Pre/Post- Deployment Evaluation}
\label{sec:pre_post_evaluation}
After creating a new HP, the off-policy evaluation (OPE) is used to evaluate the new policy before the deployment. Table~\ref{tab:ope_metrics} shows a summary of main metrics reported for each data segment (here, domain-intent of the top NLU interpretation) by the OPE analysis. In the pre-deployment evaluation, a set of expert-defined guard-rails is applied to the evaluation results to ensure robust model updates, especially for business-critical cases. If a new model fails the guard-rail conditions, the deployment will be aborted, and a human expert is tasked to investigate the issue. Otherwise, the self-learning loop will continue to optimize the policy behavior incrementally based on the user feedback, as new models are trained and deployed automatically. This automated process effectively unblocks the self-learning system from the high turn-around times required for unnecessary human intervention or A/B experimentation.

\begin{table}[t]
\centering
\resizebox{\columnwidth}{!}{
\renewcommand{\arraystretch}{1.3}
\begin{tabular}{ll}
\toprule
\textbf{Metric} & \textbf{Description} \\
\hline
Replication & rate of $\Pi_{\theta}$ making actions \\
(defect/non-defect) &  similar to $\Pi_0$\\
\hline
L1-distance & average of L1-distance \\
 & between $\Pi_0$ and $\Pi_{\theta}$\\
\hline
STD of L1-distance & std of L1-distance \\
 & between $\Pi_0$ and $\Pi_{\theta}$\\
\hline
Expected reward & IPS weighted reward for $\Pi_{\theta}$ \\
 & (counterfactual estimation)\\
\hline
Expected IPS weight & average IPS weight \\
 & (ideally equal to 1.0)\\ 
\hline
Stochastic exploration & the rate of not selecting \\
(defect/non-defect) & the highest scoring candidate \\
\hline
\end{tabular}
}
\caption{The summary of main metrics used in the pre-deployment evaluation.} 
\label{tab:ope_metrics}
\end{table}

OPE provides valuable insights about the performance of a model prior to the deployment; however, OPE estimates may suffer from an estimation bias due to weight clipping usually used to bound the IPS weights and high variance due to the log dataset coverage issues~\citep{swaminathan2016off,joachims2018deep,sachdeva2020off}. Therefore, it is essential to track the post-deployment performance of deployed polices and measure the empirical replication and user experience metrics.

\section{Experiments}
\subsection{Setup}
To evaluate the proposed self-learning skill routing method, we conducted extensive online and offline experiments in real-world production settings. In this section, we use the term baseline to refer to an implementation of a policy similar to the relabeling approach suggested by~\citet{park2020scalable}. 

We conducted online A/B experiments involving about 6M unique customers where the baseline policy served the control, and the self-learning models served treatment customers.
We trained four consecutive self-learning HPs (denoted by HP1 to HP4) with the cadence of one HP per week. Each model was trained on a traffic window of two weeks of treatment data, except the first treatment model (HP1) which was trained on logged data from the baseline collected prior to the experiment.
Due to A/B slot availability limitations in production, we decided not to update the RP in this A/B experiment and used OPE analysis to evaluate the performance of trained RP models. Therefore, we used a fixed RP model that replicates the baseline policy and focused on updating LPs throughout the experiment. We set the desired level of minimum replication for individual intents ($\tilde{\kappa}$) to $90\%$.

Additionally, we had an initial A/B experiment consisting of seven LP and two RP model updates over 49 days, demonstrating stable, steady improvements over a long period of time.
However, due to certain deployment issues, the schedule of model updates was impacted and we decided to present those results in the appendix.

\subsection{Results}
Figure~\ref{fig:ab_rewards} shows the percentage of the difference between the treatment and control for the A/B tested models. From the figure, the proposed self-learning model improves the average reward showing a general trend of improvement over iterations. Note that in a highly-optimized production system a $1\%$ improvement is considered a significant improvement in the user experience. Here, we use bootstrapping method with eight re-samples to find $95\%$ confidence intervals and show them with filled regions in the figure. Note that each reported value is the average of about 40M utterances collected over a week. Comparing the performance of the HP3 and HP4 models, we can see a regression with the forth model refresh that was predicted by OPE. However, since the reward regression did not exceed our pre-deployment guard-rail tolerance values, the deployment was proceeded.

\begin{figure}[t]
    \centering
        \includegraphics[width=\linewidth]{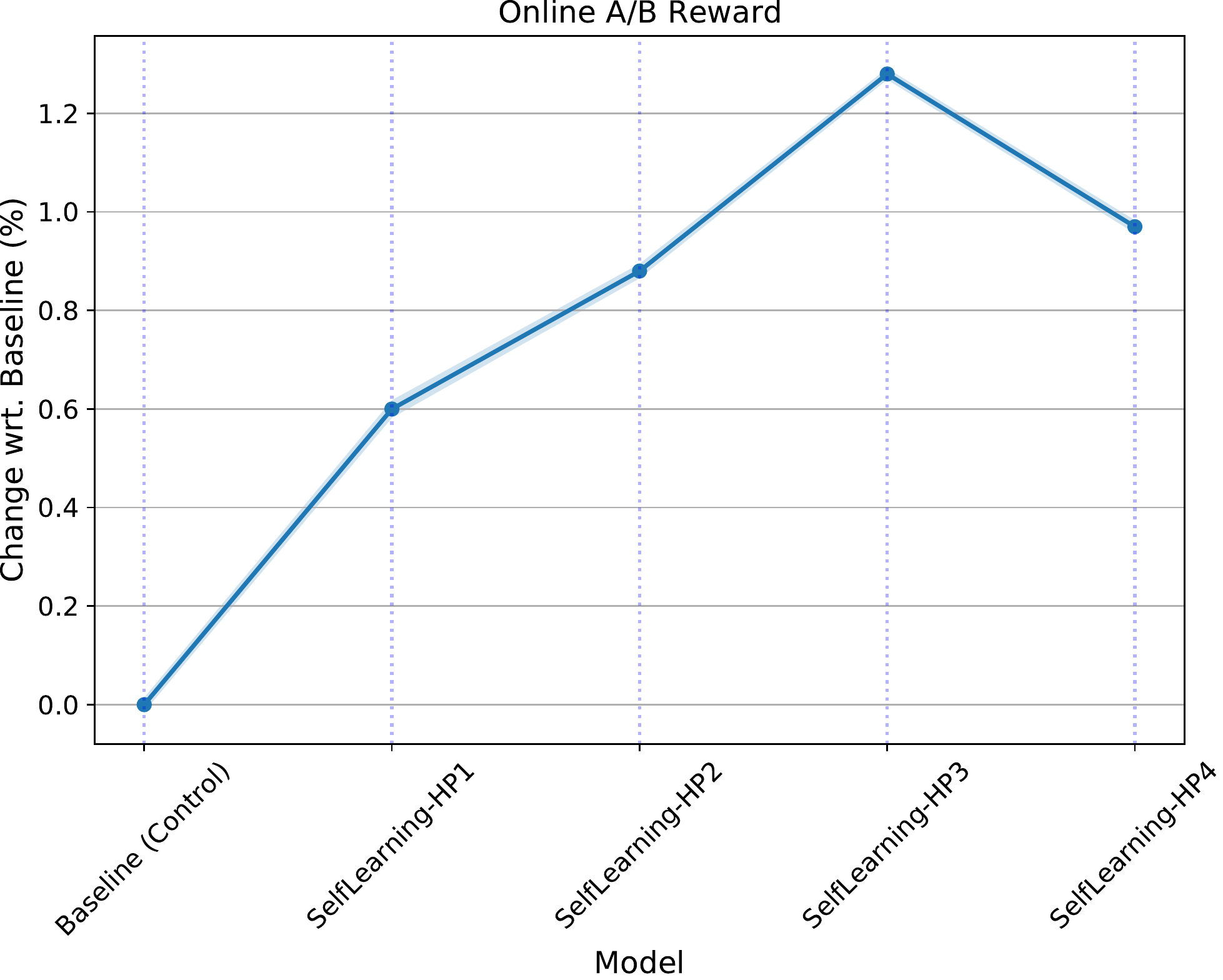}
        \caption{The comparison of the online reward measured for the baseline policy and four iterations of the self-learning model. We report the percentage of change normalized wrt. the baseline control policy.}
        \label{fig:ab_rewards}
\end{figure}

Table~\ref{tab:ope_results} shows OPE results comparing the four trained HPs. In addition to the reward, in this table, we report the rate of replication (i.e., the models making similar actions to the baseline) as well as the average L-1 distance of action propensities between each model and the baseline. Also, we report the rate at which the policy takes actions that are different from its highest-scoring action due to sampling of the softmax policy outputs. The general trend in Table~\ref{tab:ope_results} indicates that with each model refresh the new policy, on average, shows more reward and deviates more from the baseline policy. Also, the rate of stochastic exploration appears to be reduced with the consecutive updates perhaps as the model gets more confident.

\begin{table}[t]
\centering
\resizebox{\linewidth}{!}{
\renewcommand{\arraystretch}{1.3}
\begin{tabular}{lcccc}
\toprule
\textbf{Metric} & \textbf{HP1} & \textbf{HP2} & \textbf{HP3} & \textbf{HP4} \\
\hline
Reward (\%) & 93.37{\footnotesize$\pm${0.02}} & 93.41{\footnotesize$\pm${0.02}} & 93.88{\footnotesize$\pm${0.02}} & 93.75{\footnotesize$\pm${0.04}} \\
Replication (\%) & 98.06{\footnotesize$\pm${0.02}} & 98.01{\footnotesize$\pm${0.02}} & 97.75{\footnotesize$\pm${0.02}} & 97.71{\footnotesize$\pm${0.03}} \\
L-1 Distance & 3.6e-2{\footnotesize$\pm${4e-4}} & 3.7e-2{\footnotesize$\pm${3e-4}} & 4.7e-2{\footnotesize$\pm${5e-4}} & 4.5e-2{\footnotesize$\pm${5e-4}} \\
Stch. Explr. (\%) & 0.26{\footnotesize$\pm${0.01}} & 0.28{\footnotesize$\pm${0.01}} & 0.14{\footnotesize$\pm${0.00}} & 0.13{\footnotesize$\pm${0.01}} \\
\hline
\end{tabular}
}
\caption{OPE results comparing the performance of the four HP models on  the expected reward, replication rate, L-1 distance, and the rate of stochastic exploration.}
\label{tab:ope_results}
\end{table}

\begin{figure}[t]
    \centering
        \includegraphics[width=0.75\linewidth]{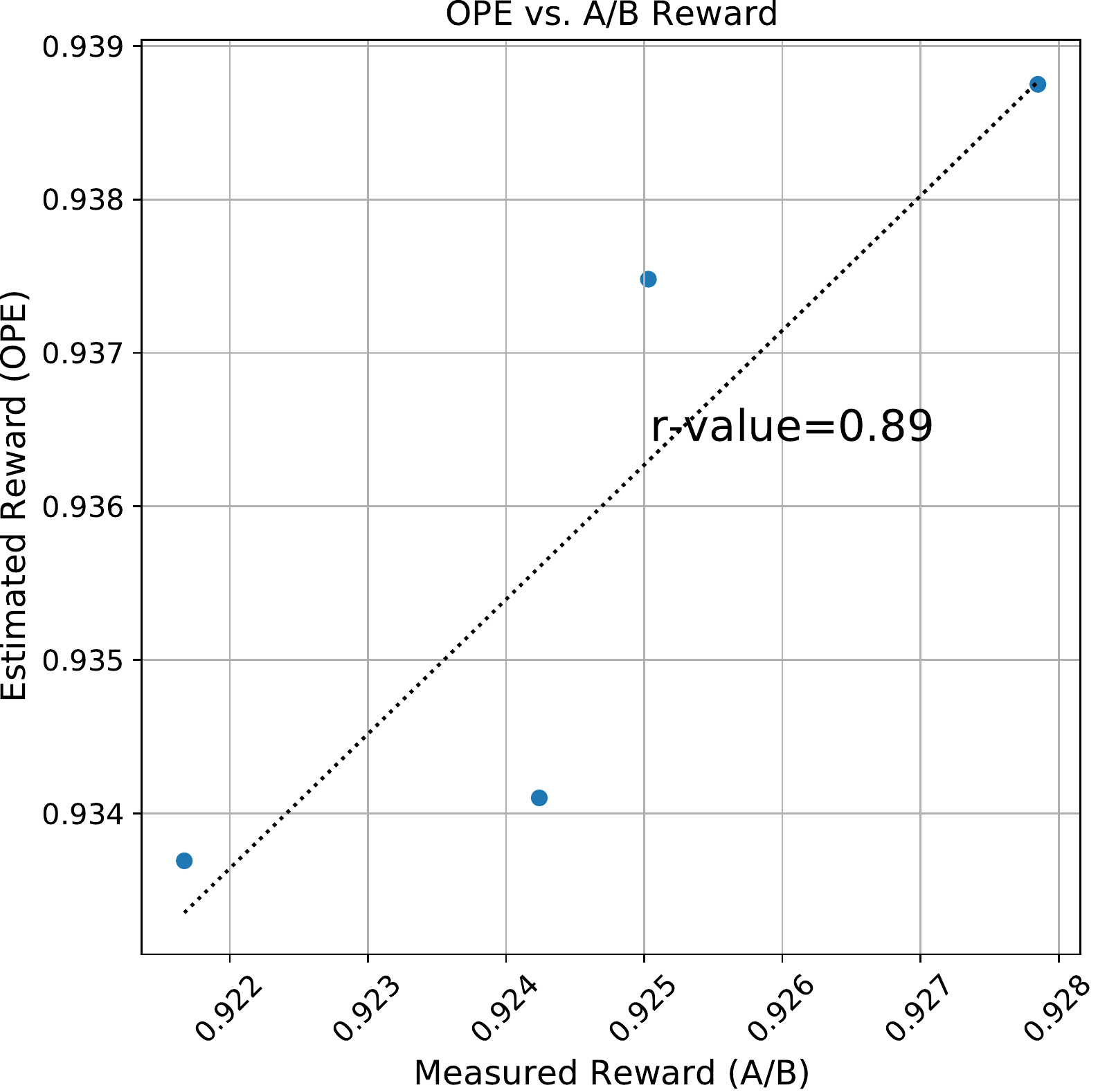}
        \caption{A calibration plot showing the correlation between the OPE reward estimates and online A/B reward measurements.}
        \label{fig:ope_calib}
\end{figure}

Figure~\ref{fig:ope_calib} compares the empirically measured reward values using online A/B experiments with OPE estimates. From the calibration plot, the OPE estimates tend to have different absolute values but show a high correlation (r-value=0.89) compared to the empirical measurements. Accordingly, OPE is capable of providing insight into how the performance of a new model would compare to the current model if we were to deploy the new model. 

\begin{figure}[t]
    \centering
        \includegraphics[width=0.80\linewidth]{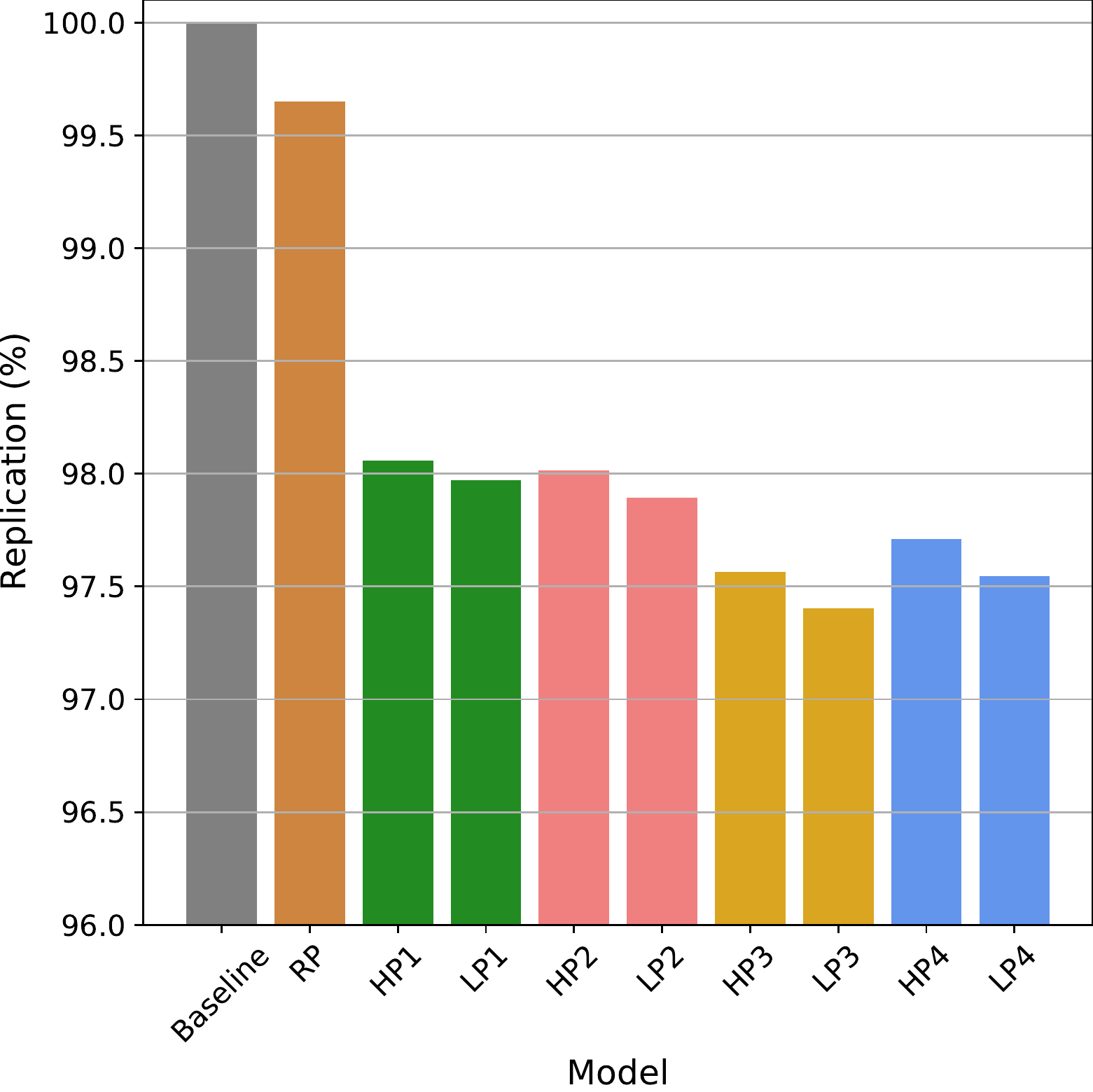}
        \caption{The comparison of the replication rates with respect to the baseline policy for the trained RP, LP, and HP models.}
        \label{fig:hp_lp_replication}
\end{figure}

In Figure~\ref{fig:hp_lp_replication}, we compare the replication rates with respect to the baseline policy for the trained RP, LP, and HP models. From this result, RP shows very high replication rates. When comparing the HP and LP replication rates, we can see HP shows a higher replication rate as the RPDR logic is adjusting the replication rate for individual intents to be no less than the desired threshold.

In addition to the presented quantitative results, we present a qualitative comparison of the baseline and self-learning models in the appendix.

\section{Conclusion}
We presented a novel self-learning approach for the skill routing problem in large-scale conversational AI systems. It leverages the user satisfaction signal to constantly improve routing decisions while maintaining frequent robust policy updates via a hybrid architecture and extensive offline analysis. The suggested hybrid architecture provides a fine-grained balance of replication and policy improvement for each NLU intent providing controlled model updates, especially for business-critical use-cases. We demonstrated the effectiveness of the proposed approach using extensive offline and online experiments in a commercial conversational system.

\bibliography{refs}

\begin{thebibliography}{14}
\expandafter\ifx\csname natexlab\endcsname\relax\def\natexlab#1{#1}\fi

\bibitem[{Agostaro et~al.(2005)Agostaro, Augello, Pilato, Vassallo, and
  Gaglio}]{agostaro2005conversational}
Francesco Agostaro, Agnese Augello, Giovanni Pilato, Giorgio Vassallo, and
  Salvatore Gaglio. 2005.
\newblock A conversational agent based on a conceptual interpretation of a data
  driven semantic space.
\newblock In \emph{Congress of the Italian Association for Artificial
  Intelligence}, pages 381--392. Springer.

\bibitem[{Bodigutla et~al.(2019)Bodigutla, Polymenakos, and
  Matsoukas}]{bodigutla2019multi}
Praveen~Kumar Bodigutla, Lazaros Polymenakos, and Spyros Matsoukas. 2019.
\newblock Multi-domain conversation quality evaluation via user satisfaction
  estimation.
\newblock \emph{arXiv preprint arXiv:1911.08567}.

\bibitem[{Jadhav and Thorat(2020)}]{jadhav2020towards}
Komal~P Jadhav and Sandeep~A Thorat. 2020.
\newblock Towards designing conversational agent systems.
\newblock In \emph{Computing in Engineering and Technology}, pages 533--542.
  Springer.

\bibitem[{Jiang et~al.(2015)Jiang, Hassan~Awadallah, Jones, Ozertem, Zitouni,
  Gurunath~Kulkarni, and Khan}]{jiang2015automatic}
Jiepu Jiang, Ahmed Hassan~Awadallah, Rosie Jones, Umut Ozertem, Imed Zitouni,
  Ranjitha Gurunath~Kulkarni, and Omar~Zia Khan. 2015.
\newblock Automatic online evaluation of intelligent assistants.
\newblock In \emph{Proceedings of the 24th International Conference on World
  Wide Web}, pages 506--516.

\bibitem[{Joachims et~al.(2018)Joachims, Swaminathan, and
  de~Rijke}]{joachims2018deep}
Thorsten Joachims, Adith Swaminathan, and Maarten de~Rijke. 2018.
\newblock Deep learning with logged bandit feedback.
\newblock In \emph{International Conference on Learning Representations}.

\bibitem[{Kachuee et~al.(2021)Kachuee, Yuan, Kim, and Lee}]{kachuee2021self}
Mohammad Kachuee, Hao Yuan, Young-Bum Kim, and Sungjin Lee. 2021.
\newblock Self-supervised contrastive learning for efficient user satisfaction
  prediction in conversational agents.
\newblock In \emph{Proceedings of the 2021 Conference of the North American
  Chapter of the Association for Computational Linguistics: Human Language
  Technologies}, pages 4053--4064.

\bibitem[{Karampatziakis et~al.(2019)Karampatziakis, Kochman, Huang, Mineiro,
  Osborne, and Chen}]{karampatziakis2019lessons}
Nikos Karampatziakis, Sebastian Kochman, Jade Huang, Paul Mineiro, Kathy
  Osborne, and Weizhu Chen. 2019.
\newblock Lessons from contextual bandit learning in a customer support bot.
\newblock \emph{arXiv preprint arXiv:1905.02219}.

\bibitem[{Li et~al.(2021)Li, Park, Dara, Nam, Lee, Kim, Matsoukas, and
  Sarikaya}]{li2021neural}
Han Li, Sunghyun Park, Aswarth Dara, Jinseok Nam, Sungjin Lee, Young-Bum Kim,
  Spyros Matsoukas, and Ruhi Sarikaya. 2021.
\newblock Neural model robustness for skill routing in large-scale
  conversational ai systems: A design choice exploration.
\newblock \emph{arXiv preprint arXiv:2103.03373}.

\bibitem[{Park et~al.(2020{\natexlab{a}})Park, Yuan, Kim, Zhang, Spyros, Kim,
  Sarikaya, Guo, Ling, Quinn et~al.}]{park2020large}
Dookun Park, Hao Yuan, Dongmin Kim, Yinglei Zhang, Matsoukas Spyros, Young-Bum
  Kim, Ruhi Sarikaya, Edward Guo, Yuan Ling, Kevin Quinn, et~al.
  2020{\natexlab{a}}.
\newblock Large-scale hybrid approach for predicting user satisfaction with
  conversational agents.
\newblock \emph{arXiv preprint arXiv:2006.07113}.

\bibitem[{Park et~al.(2020{\natexlab{b}})Park, Li, Patel, Mudgal, Lee, Kim,
  Matsoukas, and Sarikaya}]{park2020scalable}
Sunghyun Park, Han Li, Ameen Patel, Sidharth Mudgal, Sungjin Lee, Young-Bum
  Kim, Spyros Matsoukas, and Ruhi Sarikaya. 2020{\natexlab{b}}.
\newblock A scalable framework for learning from implicit user feedback to
  improve natural language understanding in large-scale conversational ai
  systems.
\newblock \emph{arXiv preprint arXiv:2010.12251}.

\bibitem[{Sachdeva et~al.(2020)Sachdeva, Su, and Joachims}]{sachdeva2020off}
Noveen Sachdeva, Yi~Su, and Thorsten Joachims. 2020.
\newblock Off-policy bandits with deficient support.
\newblock In \emph{Proceedings of the 26th ACM SIGKDD International Conference
  on Knowledge Discovery \& Data Mining}, pages 965--975.

\bibitem[{Sammut(2001)}]{sammut2001managing}
Claude Sammut. 2001.
\newblock Managing context in a conversational agent.
\newblock \emph{Linkoping Electronic Articles in Computer \& Information
  Science}, 3(7).

\bibitem[{Sarikaya(2017)}]{sarikaya2017technology}
Ruhi Sarikaya. 2017.
\newblock The technology behind personal digital assistants: An overview of the
  system architecture and key components.
\newblock \emph{IEEE Signal Processing Magazine}, 34(1):67--81.

\bibitem[{Swaminathan et~al.(2016)Swaminathan, Krishnamurthy, Agarwal,
  Dud{\'\i}k, Langford, Jose, and Zitouni}]{swaminathan2016off}
Adith Swaminathan, Akshay Krishnamurthy, Alekh Agarwal, Miroslav Dud{\'\i}k,
  John Langford, Damien Jose, and Imed Zitouni. 2016.
\newblock Off-policy evaluation for slate recommendation.
\newblock \emph{arXiv preprint arXiv:1605.04812}.

\end{thebibliography}
\bibliographystyle{./template/acl_natbib}

\clearpage
\appendix
\section{Appendix}

\subsection{Qualitative Results}

Table~\ref{tab:ab_qualitative} shows a qualitative comparison of the baseline (relabeling approach) and self-learning (bandit-based HP) decisions. We provide the actual user utterance transcription and the selected skill using each method. The green color shows the skills providing the best user experience. 

\begin{table*}[b]
\centering
\resizebox{0.85\linewidth}{!}{
\renewcommand{\arraystretch}{1.3}
\begin{tabular}{llll}
\toprule
 & & \multicolumn{2}{c}{\textbf{Selected Skill}} \\
\textbf{Example} & \textbf{Utterance} & \textbf{Baseline Model} & \textbf{Self-Learning Model} \\
\hline
win-1 & \textit{what is the best seasoning for mahi-mahi} & shopping & \textcolor{ForestGreen}{knowledge (Q\&A)} \\ 
win-2 & \textit{show me wildlife photography} & shopping & \textcolor{ForestGreen}{photos (gallery)} \\ 
win-3 &\textit{give me n. b. c. news} & knowledge (Q\&A) & \textcolor{ForestGreen}{daily briefing (news)} \\ 
win-4 &\textit{get some cheeto puffs} & knowledge (Q\&A) & \textcolor{ForestGreen}{shopping} \\ 
win-5 & \textit{set up [DEVICE NAME]} & pairing (Bluetooth) & \textcolor{ForestGreen}{setup (home automation) } \\ 
loss-1 &\textit{what is the best song in the world} & \textcolor{ForestGreen}{knowledge (Q\&A)} & find music \\
loss-2 & \textit{play announcement} & \textcolor{ForestGreen}{announcement} & get message \\
\hline
\end{tabular}
}
\caption{A few examples of skill routing for the baseline and self-learning models. The green color is used to indicate skills providing the best user experience.}
\label{tab:ab_qualitative}
\end{table*}

\subsection{Additional A/B Experiment Results}
\begin{figure*}[b]
    \centering
        \includegraphics[width=0.75\linewidth]{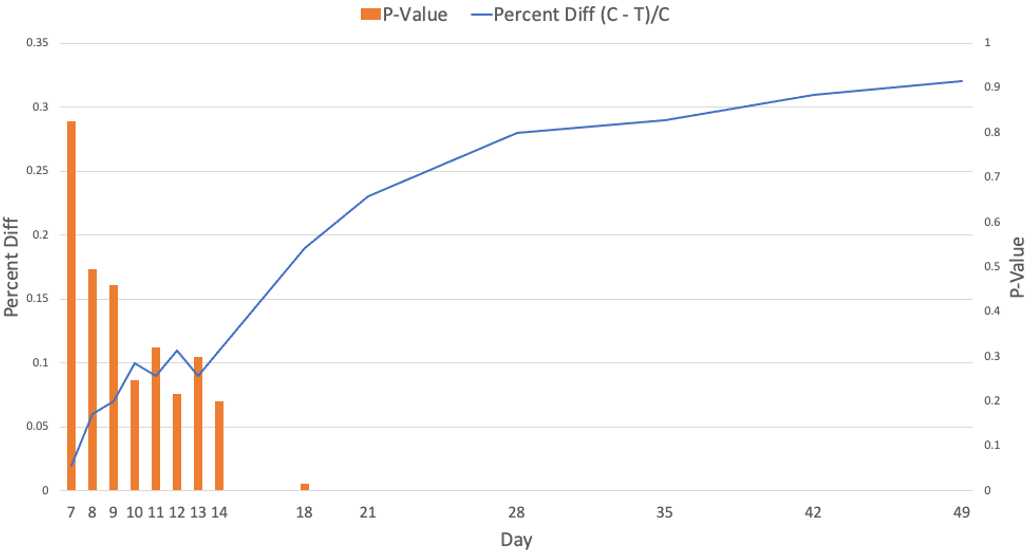}
        \caption{The percentage of difference for the measured reward between the control (relabeling baseline) and treatment (self-learning) slots over a 49-day initial proof-of-concept A/B experiment.}
        \label{fig:ab_results_old}
\end{figure*}

Figure~\ref{fig:ab_results_old} shows the trend of change in the reduction of user dissatisfaction rate over a 49-day long A/B experiment. During the A/B experiment, we updated the LP model seven times and the RP model two times. As this long-running A/B was one of our initial proof-of-concept experiments on the production system, we faced several deployment and technical issues that impacted the schedule of LP and RP updates. Nonetheless, from the results, we can see consistent and statistically significant improvements in user satisfaction.

\end{document}